\documentclass[10pt,twocolumn,letterpaper]{article}

\usepackage{cvpr}
\usepackage{times}
\usepackage{epsfig}
\usepackage{graphicx}
\usepackage{amsmath}
\usepackage{amssymb}

\usepackage{algpseudocode}
\usepackage{algorithm}
\usepackage{subfigure}

\usepackage[pagebackref=true,breaklinks=true,letterpaper=true,colorlinks,bookmarks=false]{hyperref}

\cvprfinalcopy 

\def\cvprPaperID{0000} 

\ifcvprfinal\fi

\usepackage{graphicx}
\usepackage{amsmath,amssymb} 
\usepackage{color}
\usepackage{array,multirow}

\begin{document}

\title{Deep unsupervised 3D human body reconstruction from a sparse set of landmarks\thanks{This is a preprint of an article published in International Journal of Computer Vision. The final authenticated version is available online at: https://doi.org/10.1007/s11263-021-01488-2}}

\author{\parbox{16cm}{\centering
    {\large Meysam Madadi$^{1}$, Hugo Bertiche$^{2}$ and Sergio Escalera$^{1,2}$}\\
    {\normalsize
    $^1$ Computer Vision Center, Edifici O, Campus UAB, 08193 Bellaterra (Barcelona), Catalonia Spain\\
    $^2$ Dept. Mathematics and Informatics, Universitat de Barcelona, Catalonia, Spain}}
}

\maketitle

\begin{figure*}[!ht]
    \centering
    \includegraphics[width=.9\textwidth]{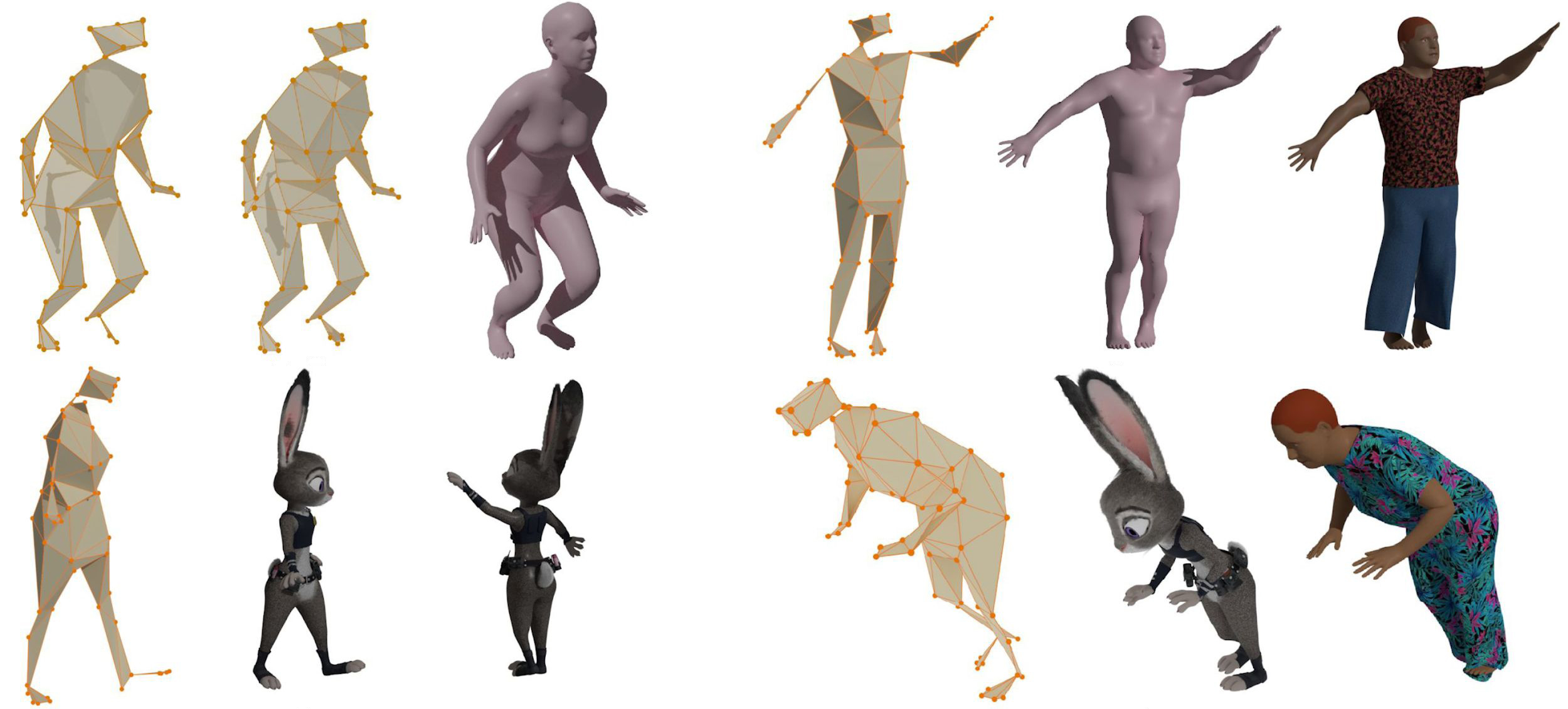}
    \caption{Applications of landmarks to body surface reconstruction. The input to the system is a sparse set of landmarks. a) Missing landmarks in the input are recovered and then body surface is reconstructed. b) Garments are simulated on top of the body surface and rendered with textures. c) Any subject can be reconstructed and re-posed from estimated pose and shape parameters. d) More examples of retargeting and garment simulation. The edges are added to the landmarks for visualization purposes.}
    \label{fig:motivation}
\end{figure*}

\begin{abstract}
In this paper we propose the first deep unsupervised approach in human body reconstruction to estimate body surface from a sparse set of landmarks, so called \textit{DeepMurf}. We apply a denoising autoencoder to estimate missing landmarks. Then we apply an attention model to estimate body joints from landmarks. Finally, a cascading network is applied to regress parameters of a statistical generative model that reconstructs body. Our set of proposed loss functions allows us to train the network in an unsupervised way. Results on four public datasets show that our approach accurately reconstructs the human body from real world mocap data.
\end{abstract}

\section{Introduction}
Reconstruction of 3D human body has a great applicability in many domains including pose and shape retargeting, movie editing, videogame industry and virtual reality, just to name a few. This is a particularly challenging problem, since the solution must deal with 3D joint locations and orientations along with subject specific body surface. Besides, data acquisition and annotation is an expensive process, specially for supervised approaches.

Given the difficulty of acquiring 3D ground truth body surface, this problem is tackled unsupervisedly in different ways in the literature. On the one hand, image-based approaches \cite{kanazawa2018end,omran2018neural} try to reconstruct 3D body from 2D data gathered from images. However, this is known as ill-posed since depth is lost in the projection to the image plane and 2D-to-3D reconstruction can have multiple solutions for the same 2D data. Some authors propose multi-camera setups \cite{joo2018total,rhodin2016general,mehrizi2018toward} to cope with this problem. On the other hand, mocap-based pose estimation has become a standard procedure in domain specific applications like movie and videogame industry. However, standard procedures are not able to reconstruct body surface. In this regard, Loper \textit{etal.} \cite{loper2014mosh} showed that a sparse set of physical landmarks attached to the body is enough to reconstruct the whole body surface. 

In this paper we focus on mocap-based solutions. Prior works \cite{loper2014mosh,mahmood2019amass} are based on regular optimization techniques in which a statistical body model is fit to the input landmarks. These approaches are able to reconstruct the body surface accurately, though, in a significant amount of time. Besides, optimization is applied in several steps, e.g., body shape is optimized first and later pose is conditioned on shape. This further prevents these approaches to be utilized in on-the-fly applications. On the contrary, in this paper we propose a deep unsupervised approach to reconstruct 3D body surface from single frame mocap data which is fast in training and testing time, able to generate accurate results. Working on single frames, rather than sequences, relaxes the network from the need of enormous temporal data, allowing our approach to be trained on small custom datasets. To the best of our knowledge this is the first time a deep unsupervised approach is applied to this problem.

Specifically, we build our approach based on a widely used statistical generative model (SMPL) \cite{loper2015smpl}. SMPL receives body pose and shape parameters, updates a template mesh and generates body surface through forward kinematic. The goal of this paper is to estimate SMPL pose and shape parameters from mocap data through deep learning such that generated surface best fits the input 3D landmarks. There are several challenges in this task: 1) landmarks are noisy and sometimes missing, 2) although SMPL is differentiable, it is sensitive to noise, and 3) the space of human pose and shape is highly non-linear. These challenges produce many local minima and an efficient architecture and training procedure are crucial for model generalization. Given the aforementioned challenges, one goal in this paper is to design a network that can work well in small size custom datasets.

We inspire our architecture from SMPLR \cite{madadi2018smplr}, in which a set of landmarks are predicted from input RGB images and used to estimate body pose and shape. SMPLR is a supervised approach which is not applicable on the problem at hand. Our main contribution is to redesign SMPLR for unsupervised training. Specifically, our contributions are as follows. Our proposed architecture is composed of a denoising autoencoder (similar to SMPLR) to recover missing landmarks, an attention model to estimate joints from landmarks, and a cascading regression model to estimate body pose and shape parameters. We train the model unsupervisedly, that is, 3D joints and body pose and shape parameters are unknown. To do so we propose several loss functions including regularization on pose and shape parameters, landmarks-to-surface loss and denosing autoencoder and attention model loss functions. We also propose a novel unposing layer which helps to generalize better when there is a low amount of data. Finally, we provide an extensive analysis of the architecture and loss functions on four public datasets \cite{varol2017surreal,mahmood2019amass,hoyet12hand,cmu2001}. Particularly, our approach can deal with missing landmarks, accurately estimates joints and body surface (including hands) and performs well when trained on small datasets.

\section{Related works}

Human pose and shape recovery has been an extensively studied field of research in the recent years. There is a large literature on the topic which could be classified according to considered input data: image, IMUs, mocap or combinations, or according to methodology: energy optimization, database search or deep learning. Regarding deep learning, we could further subdivide it into supervised and unsupervised learning. Nonetheless, to the best of our knowledge, we are the first ones to try a direct mapping from sparse mocap sensor data to body surface through unsupervised deep learning.

\subsection{Modalities}
\textbf{Image.} This category is the most extensive. On one hand, we have multi-view setups  \cite{joo2018total,rhodin2016general,mehrizi2018toward}, which can be tackled through energy optimization or deep learning. Multi-view data significantly reduces problem complexity, but it requires a constrained scenario, which limits applicability. On the other hand, we find monocular RGB approaches. Deep learning methodologies are  predominant in current literature, outperforming traditional, non-deep, strategies. 
Volumetric body prediction is a highly complex task, so most works rely on parametric models such as SMPL \cite{loper2015smpl} to predict 3D body model. This is done by direct regression \cite{kanazawa2018end} or through intermediate representations \cite{pavlakos2018learning,omran2018neural,madadi2018smplr,varol2018bodynet,XNect_SIGGRAPH2020}. We also find works that rely on depth maps \cite{bogo2015detailed,achilles2016patient}, though, as with multi-view, it simplifies the problem while requiring a specific setup. Works based on RGB data are prone to be domain-dependant, impairing their generalization and applicability. In this work we propose recovering body pose and shape only from mocap-like data which yields a domain-independant model.

\textbf{IMUs.} Inertial Measurement Units are a very common sensor, found in smartphones, gaming devices and airplanes. They provide acceleration and orientation data, but no location. Due to their availability, many researchers proposed methodologies to obtain body pose from sparse IMUs. Authors of \cite{von2017sparse} propose an energy optimization approach to obtain SMPL parameters from temporal IMU data. These works \cite{slyper2008action,tautges2011motion} compare input data against a prerecorded database. Schwarz et al. \cite{schwarz2009discriminative} propose a Gaussian Process Regression to map IMU data to full body pose, though, it has generalization problems. Finally, Huang et al. \cite{huang2018deep} apply deep learning to the recordings of 6 IMU sensors to predict body pose with a RNN. While IMUs are cheap, their lack of location data renders them useless to determine subject body shape and often temporal data is required to solve pose ambiguities. These drawbacks are not present for mocap sensors.

\textbf{Mocap.} This data contains the location of a set of sparse landmarks evenly placed through body surface. They are the film and animation industry standards on motion capture because of the accuracy of measurements, while on the other hand, require an specific multi-view setup for correct location tracking. Similar to IMU-based approaches, we find energy optimization methods, such as MoSH \cite{loper2014mosh} or \cite{park2006capturing}, where body pose and shape are obtained from sparse mocap markers. Although energy optimization methods generate good results, they do not achieve real-time performance. Other approaches have also been explored \cite{chai2005performance,liu2011realtime}, where input data is compared against pre-recorded databases. These approaches mainly suffer from lack of generalization. Instead of sparse landmarks, some works \cite{groueix2018b,prokudin2019efficient,bhatnagar2020combining,bhatnagar2020loopreg} predict body (or outfit) surface from dense point clouds. These approaches perform as registration techniques to find correspondences. Our work proposes a mapping from sparse mocap data to SMPL pose and shape parameters through deep learning, which, to the best of our knowledge, has not been previously explored.

\textbf{RGB+IMU/Mocap.} Some works like \cite{von2016human,von2018recovering} use combination of modalities, RGB plus IMUs in this case, to improve accuracy on pose prediction by complementing the drawbacks of each data type with the other. Similarly, in \cite{trumble2017total}, Trumble et al. propose an RGB multi-view plus IMUs setup, from which volumetric probabilistic
visual hull data is extracted and fed to a 3DCNN for human 3D pose prediction. In \cite{zhao2012combining}, Zhao et al. use markers (mocap-like data) and kinect sensor data (RGB-D) to obtain accurate 3D hand models as an optimization problem.

\subsection{Unsupervised approaches} 
Aforementioned deep-based strategies work in a supervised scenario. Unsupervised learning has not been widely explored yet in human pose and shape recovery. As proposed in \cite{kudo2018unsupervised,chen2019unsupervised}, 3D pose can be mapped from RGB in an unsupervised manner only after decomposing the problem in 2D joint detection from an image and 2D-to-3D joint lifting. The 2D joints are assumed to be accurate enough and the lifting part is learnt without direct supervision by back-projecting predictions and ensuring consistency w.r.t. 2D \textit{annotations}. Human shape recovery without supervision has not been tackled yet, but in \cite{insafutdinov2018unsupervised}, Insafutdinov et al. propose an unsupervised learning of an RGB-to-3D mapping for generic objects, also based on back-projection and comparison w.r.t. visual evidence.

\begin{figure*}[!ht]
    \centering
    \includegraphics[width=\textwidth]{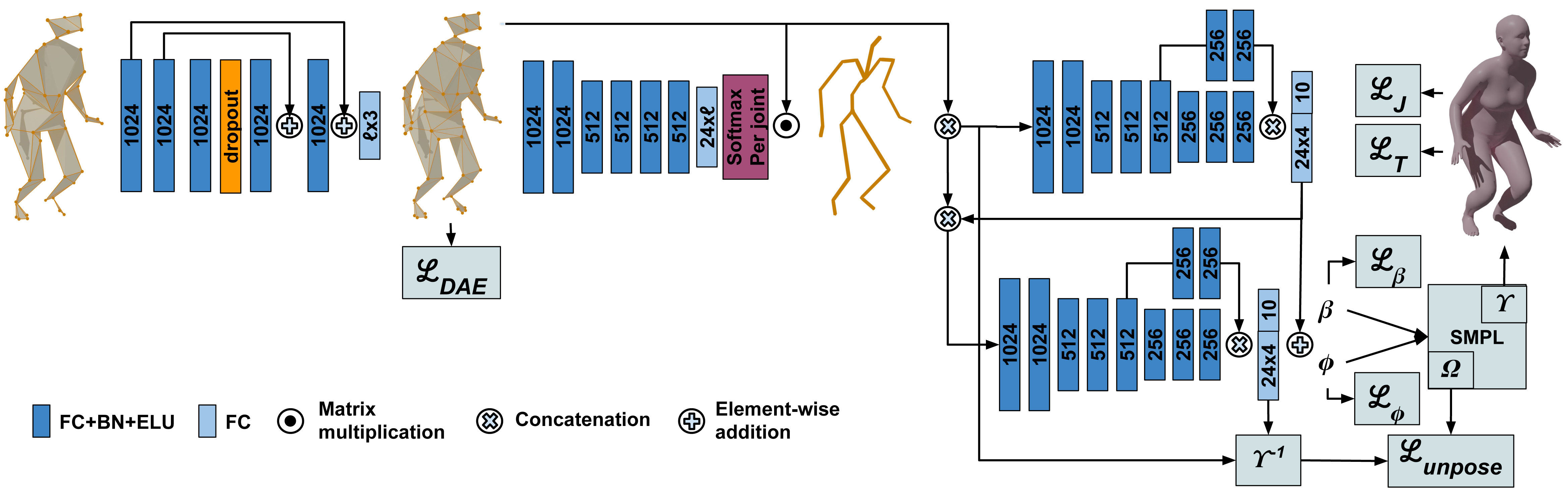}
    \caption{Architecture pipeline. First missing landmarks are recovered in denoising autoencoder network. Then the attention network is used to estimate body joints. Both landmarks and body joints are fed to the cascading network which regress shape ($\beta$) and pose ($\phi$ in terms of quaternion) parameters. Finally, the body surface is generated by SMPL. Our proposed loss functions allow us to train the network unsupervisedly.}
    \label{fig:pipeline}
\end{figure*}

\section{Unsupervised surface reconstruction from landmarks}
Let $\mathcal{X}=\{\mathbf{L}\}_{i=1}^{n}$ be a dataset where $\mathbf{L}_i \in \mathbb{R}^{l\times3}$ is a set of $l$ landmarks in 3D coordinates for a given frame. Landmarks have several properties: 1) the set $\mathbf{L}$ is ordered, that is, $j-$th landmark is always attached to the same location on the body, 2) landmarks are sparse, i.e. they cover a few locations on the body ($l<100$ in this paper), and 3) landmarks are noisy with missing data. There are two types of noise in the landmarks: 1) small perturbation in the attachment location and 2) large measurement error that may cause some landmarks appear far from where they should be. Finally, there is no guarantee that there will always be $l$ landmarks attached to the body. Therefore, $l$ is the number of all possible landmarks in the dataset $\mathcal{X}$. In this paper we use the same set of landmarks as in \cite{loper2014mosh} and assume missing or corrupted landmarks are known beforehand. We define valid landmarks as a mask by matrix $\mathbf{M} \in \{0,1\}^{l\times3}$.

The goal is to train a neural network on the dataset $\mathcal{X}$ to output a dense set of 3D body surface points $\mathbf{T}_{out}\in\mathbb{R}^{p\times3}$ that best fits the data. To do so, we apply a hybrid approach which combines deep learning with SMPL, a statistical generative model, that reconstructs body surface. SMPL receives axis-angle pose parameters $\theta \in \mathbb{R}^{24\times3}$ along with shape coefficients $\beta \in \mathbb{R}^{10}$ to generate body surface $\mathbf{T}_{out}$ with $p=6890$. Therefore, the network is simplified to regress SMPL parameters from input landmarks. Regressing SMPL parameters allows us to apply pose and shape retargeting in custom applications, as shown in Fig. \ref{fig:motivation}. The original SMPL implementation has two separate models for male and female. As a common practice, we use a neutral gender model that simplifies the training.

The pipeline of the architecture is shown in Fig. \ref{fig:pipeline}. First, we apply a denoising autoencoder to recover missing and noisy landmarks. Then an attention network is used to estimate body joints. Finally, recovered landmarks and estimated joints are concatenated as input to a cascading network to regress SMPL pose and shape parameters. Next, we explain details of each part and how we train the network unsupervisedly.

\subsection{Recover missing landmarks}
Missing landmarks are unavoidable during the setup and capturing process. Therefore a neural network must be able to deal with a variable number of landmarks per frame (or per sequence of frames). A standard neural network (e.g. a MLP network) requires ordered and fixed size arrays. By knowing landmark labels and matrix $\textbf{M}$, we can fill valid landmarks in each frame to form the input matrix $\mathbf{L}$. Then missing or erroneous landmarks are set to zero.

Although neural networks are able to handle data arrays with few zero values, there is no guarantee they implicitly learn the patterns of missing information. It has been shown that denoising autoencoders \cite{vincent2010stacked} are useful tools to learn local representations of corrupted data. Therefore, given a large dataset, it is possible to estimate a missing landmark in a frame from neighbor frames in the representation space. In this paper we apply a denoising autoencoder network (DAE, as proposed in \cite{madadi2018smplr}) by the usage of fully connected layers, dropout and skip connections, as shown in Fig. \ref{fig:pipeline}. We train DAE for one frame using L1 loss as:
\begin{equation}
    \mathcal{L}_{DAE}=\frac{1}{l \times 3} \sum_{j=1}^{l\times3} \mathbf{M}_j | \mathbf{L}_j - \hat{\mathbf{L}}_j |,
\end{equation}
where $\hat{\mathbf{L}}$ is the output of DAE as the set of estimated landmarks. Note that we normalize $\mathbf{L}$ beforehand by subtracting its mean value. Therefore, we make $\mathbf{L}$ translation invariant.

\subsection{Regress pose and shape parameters}
Our goal is to estimate SMPL pose and shape parameters from landmarks. SMPL pose is defined by axis-angle rotation of each joint w.r.t. its parent joint in the skeleton kinematic tree. Therefore, SMPL pose is a combination of local representations for each joint. This is while the landmark coordinates are represented globally. We believe a standard MLP network has difficulties to implicitly learn a direct mapping from these global landmarks to local relative axis-angle rotations, as we show in the experiments. This mapping suffers from many local minima regardless of the capacity of the network. This behavior is observed in image-based body reconstruction domain as well. Kanazawa \textit{et al.} \cite{kanazawa2018end} directly mapped images, as global representations, to SMPL parameters trying to handle local minima through adversarial training. Later, Madadi \textit{et al.} \cite{madadi2018smplr} showed that a two-step mapping could significantly improve the results, i.e. by first mapping to, easier to extract, intermediate representations and then mapping them to SMPL parameters. Similarly, we first extract body joints from landmarks as complementary information since they are basis coordinates for these relative rotations, and use them along with landmarks to predict SMPL parameters. We do this process unsupervisedly in a unified pipeline. However, one can apply standard mocap pose estimators for this task. 

Fortunately, body joint locations can be interpolated from surface vertices. In the case of sparse landmarks, the accuracy of interpolated joints depends on the placement of the landmarks. We use the same set of landmarks as in \cite{loper2014mosh}, where an optimization is applied for the placement and importance of landmarks. It is also a standard procedure in mocap data to place at least one landmark around main body joints (e.g. wrist, elbow, hip, knee and ankle). We found standard mocap landmarks are rich enough to approximate the joints. To do so, we design an attention network (ATN) which receives updated landmarks $\hat{\mathbf{L}}=\mathbf{M}\odot\mathbf{L}+(1-\mathbf{M})\odot\hat{\mathbf{L}}$ and outputs joints $\mathbf{J}_{in} \in \mathbb{R}^{m\times3}$ where $\odot$ is the hadamard product and $m=24$ is the number of SMPL joints. The architecture can be seen in Fig. \ref{fig:pipeline}. $\mathbf{J}_{in}$ is computed as:
\begin{equation}
    \mathbf{J}_{in} = \sigma(\mathbf{A}) \cdot \hat{\mathbf{L}},
\end{equation}
where $\mathbf{A} \in \mathbb{R}^{m \times l}$ is the last reshaped layer of ATN and $\sigma$ is the softmax over the second dimension of $\mathbf{A}$. Let $J_{in}^0$ be the estimated root joint location in the kinematic tree. We subtract both $\mathbf{J}_{in}$ and $\hat{\mathbf{L}}$ from $J_{in}^0$ to have translation invariant data.

Now, $\theta$ and $\beta$ can be regressed from $\mathbf{J}_{in}$ and $\hat{\mathbf{L}}$. First, we explain the modifications we apply to the pose parameters. In the SMPL pipeline, axis-angles $\theta$ are converted to rotation matrices through Rodrigues formulation. It is known that axis-angles are not unique and Rodrigues function is not one-to-one. This is problematic in the training due to its instability and convergence to wrong values. In the literature, axis-angles are replaced with rotation matrices, bypassing Rodrigues function. However, we believe this is not suitable either, because there is no guarantee the regression network yields valid rotation matrices. That means predicted matrices are not orthonormal. Instead, we propose to replace axis-angles $\theta$ with quaternions $\phi \in \mathbb{R}^{m\times4}$ which are known to be unique and can be easily converted to rotation matrices.

Any proposed network must be able to efficiently map between two highly nonlinear spaces, i.e. from $\mathbf{J}_{in}$ and $\hat{\mathbf{L}}$ to $\phi$ and $\beta$. To cope with this, we propose a cascade of sequential networks $\{\Psi_0,\Psi_1,...,\Psi_c\}$. All $\Psi$ networks have a similar architecture without sharing weights. Let $\{\phi,\beta\}=\Psi_0(\mathbf{J}_{in},\hat{\mathbf{L}};\mathbf{\omega}_0)$ be our first pose and shape regressor where $\mathbf{\omega}_0$ is its trainable parameters. Then each $\Psi_i$ , $ i\neq0,$ is defined as $\{\phi,\beta\}=\Psi_i(\mathbf{J}_{in},\hat{\mathbf{L}},\Psi_{i-1};\mathbf{\omega}_i)+\Psi_{i-1}$. The architecture with one cascade can be seen in Fig. \ref{fig:pipeline}.

\subsection{How do we train the network?}
In this section, we explain details of the applied loss functions and training procedure. We note that the only available information are $\mathbf{L}$ and $\mathbf{M}$ and the network must learn $\mathbf{J}_{in}$, $\beta$ and $\phi$ unsupervisedly.

\textbf{Regularization on $\beta$ and $\phi$.} Due to the lack of ground truth data on $\beta$ and $\phi$, we do not know the real distribution of these parameters. However, we can define upper and lower bounds on them. This is particularly important to teach the network to be aware of valid parameters, since SMPL is sensitive to noise and can converge to invalid parameters. Pose regularization for one frame is defined as:
\begin{equation}
    \mathcal{L}_{\phi} = \frac{1}{m\times4}\sum max(0, \mathbf{B}_l - \phi) + max(0, \phi - \mathbf{B}_u),
\end{equation}
where $\mathbf{B}_l,\mathbf{B}_u \in \mathbb{R}^{m\times4}$ are pose lower and upper bounds. We set $\mathbf{B}_l$ and $\mathbf{B}_u$ manually by checking valid angles of each joint in SMPL and converting them to quaternions. For shape reqularization, we apply a standard $L1$ norm to force shape parameters close to zero, as well as keeping a hard boundary on shape: 
\begin{equation}
    \mathcal{L}_{\beta} = \frac{1}{10}\sum max(0, |\beta|-5) + |\beta|.
\end{equation}

\textbf{Joints and surface loss.} Let $\mathbf{J}_{out}\in \mathbb{R}^{m\times3}$, along with $\mathbf{T}_{out}$, be SMPL outputs of joints and body surface vertices. We define a loss on $\mathbf{J}_{out}$ based on an observation: $\mathbf{J}_{in}$ error is way lower than $\mathbf{J}_{out}$. So it can be used as a teacher to $\mathbf{J}_{out}$ in the loss (computed for one frame):
\begin{equation}
    \mathcal{L}_J(\mathbf{J}_{in},\mathbf{J}_{out}) = \frac{1}{m\times3} \sum |\mathbf{J}_{in}-\mathbf{J}_{out}|.
\end{equation}
To fit the surface on landmarks one must take several challenges into account: 1) landmarks are in a distance to the surface, and 2) landmarks have perturbation in their placement on the body. To cope with the first challenge, Loper \textit{et al.} \cite{loper2014mosh} apply a loss to keep a landmark-to-surface distance higher than a threshold. We believe this loss is unnecessary as long as we update SMPL template vertices. Specifically, we add a vector of size 1cm\footnote{We find 1cm adequate for the experimented datasets, though can be adjusted for custom datasets.} to each SMPL template vertex in the direction of vertex normal. We do this once just in the training and save the network from extra complexity. To cope with the second challenge, we apply a soft landmark-to-surface assignment. That is, for each landmark we manually select and fix a patch of SMPL vertices that the landmark may appear in. Then, the nearest vertex in the patch to the landmark is the candidate for the computation of loss:
\begin{equation}
    \mathcal{L}_T(\hat{\mathbf{L}},\mathbf{T}_{out}) = \frac{1}{l\times3} \sum_{i=1}^{l} \sum_{j=1}^{3} \min_{k\in\rho_i} |\hat{\mathbf{L}}^{i,j}-\mathbf{T}_{out}^{k,j}|,
\end{equation}
where $\rho_i$ is a patch of assigned indices to $i-$th landmark. Although this loss can yield some offset error in low resolution meshes, it works well in practice and has a low complexity. 

\begin{figure*}[!ht]
    \centering
    \includegraphics[width=0.8\textwidth]{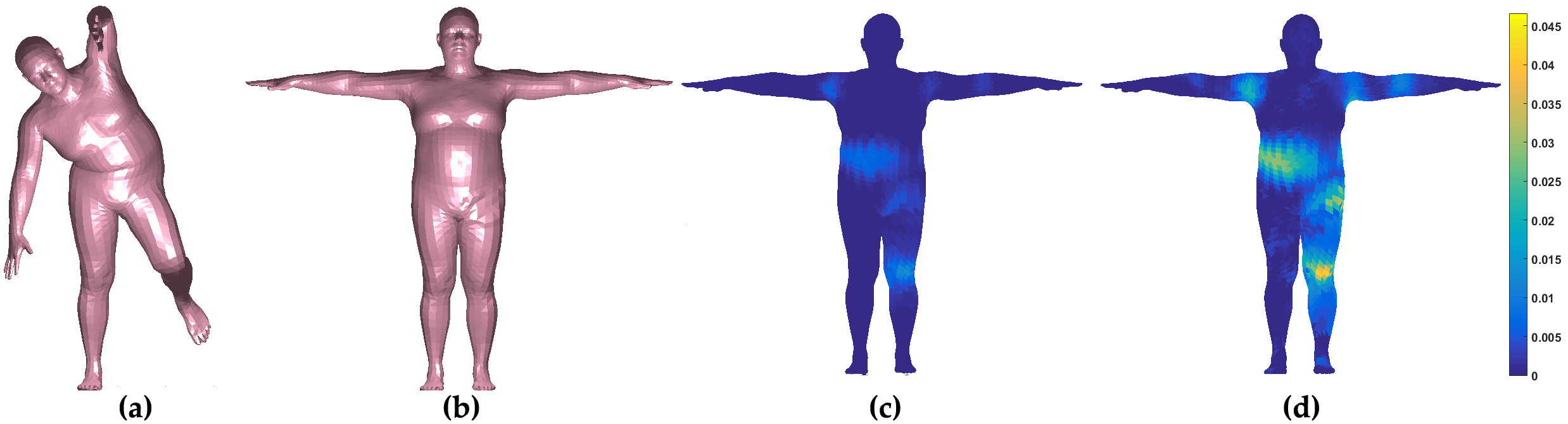}
    \caption{An example of the proposed inverse kinematic. a) A given body, b) and c) our proposed unposing and its per vertex error heatmap, and d) error heatmap for the standard recursive inverse kinematic (Eq. \ref{eq:inv_L}). }
    \label{fig:inv_res}
\end{figure*}

\textbf{Inverse kinematic loss.} SMPL is a multi-valued function, that is, there are multiple valid solutions for a given body surface. SMPL is also sensitive to noise in the loss (due to the noise in the landmarks) and it hurts back-propagated gradients. To handle this problem, we force the network to provide unique solutions for SMPL, i.e. to have a one-to-one function. To do so, we assume SMPL shaped body surface in rest pose (computed in the forward path) as a canonical surface. Then, we want joints and landmarks (i.e. $\mathbf{J}_in$ and $\hat{\mathbf{L}}$), inverted through backward SMPL, to be perfectly similar to the canonical surface. This helps the network to be aware of geometry and improves generalization. Formally, we define this loss as follows.

Let $\{\mathbf{J}_t,\mathbf{T}_t\}=\Omega(\beta,R(\phi);\mathbf{T}_t^*,\mathcal{W})$ be the SMPL function that produces shaped vertices $\mathbf{T}_t$ and joints $\mathbf{J}_t$ from template vertices $\mathbf{T}_t^*$. We note that all variables with subscript $t$ have a rest pose. $R(\phi)$ converts quaternions to rotation matrices and $\mathcal{W}$ is a precomputed set of SMPL parameters including blend shape functions and blend weights. Also, let $\mathbf{a}_{out} = \Upsilon(R(\phi),\mathbf{a}_t;\mathbf{W})$ be the forward kinematic function that transforms any given joints or landmarks $\mathbf{a}_t$ to the final posed form $\mathbf{a}_{out}$ where $\mathbf{W}\in\mathbb{R}^{6890\times m}$ is the set of blend weights. Finally, let $\Upsilon^{-1}$ be the inverse kinematic function which must be able to unpose any given joints or landmarks. Then, we define inverse kinematic loss as:

\begin{equation}
    \mathcal{L}_{unpose} = \mathcal{L}_J(\Upsilon^{-1}(R(\phi),\mathbf{J}_{in}),\mathbf{J}_t) + \mathcal{L}_T(\Upsilon^{-1}(R(\phi),\hat{\mathbf{L}}),\mathbf{T}_t).
\end{equation}

\textbf{Inverse kinematic function.} Forward kinematic function $\Upsilon$ is part of SMPL pipeline. Here, we explore the details of inverse kinematic function $\Upsilon^{-1}$. The unposing procedure is different between joints and landmarks. We first explain this process for joints.
This is done by recursive unposing of the joints in the kinematic tree through $\mathbf{R}=R^\intercal(\phi)$ where $R^\intercal$ is the transpose of rotation matrices for each joint. We show indexing operator by superscript indices, e.g. $\mathbf{J}_{in}^i$ means $i$-th joint.
\begin{equation}
    \mathbf{J}_r = \Big[\mathbf{J}_{in}^i - \mathbf{J}_{in}^{\kappa_i} \Big]_{i=1}^m,
\end{equation}
\begin{equation}
    \mathbf{G}_r = \Big[\mathbf{R}^{\kappa_i} \cdot \mathbf{G}^{\kappa_i}\Big]_{i=2}^m,
\end{equation}
\begin{equation}
    \mathbf{J}_t = \Big[\mathbf{G}_r^i \cdot \mathbf{J}_r^{i-1} + \mathbf{J}_t^{\kappa_i}\Big]_{i=2}^m.
\end{equation}
where $\kappa\in\mathbb{R}^m$ is the kinematic relationship between joints, i.e. $\kappa_i$ is the parent index of $i-$th joint. $\mathbf{G}\in\mathbb{R}^{m\times4\times4}$ is a set of $m$ transformation matrices computed from $R(\phi)$ and $\mathbf{J}_t$ (similar to SMPL). Since this procedure is recursive, $\mathbf{J}_t^1$ is set by $\mathbf{J}_{in}^1$ and $\mathbf{G}^1$ is set by an identity matrix. 

Then, we compute $\hat{\mathbf{L}}_t$ as:
\begin{equation}
    \mathbf{G}^{'} = \Big[\mathbf{R}^i \cdot \mathbf{G}^i \Big]_{i=1}^m,
\end{equation}
\begin{equation}
    \mathbf{O} = \mathbf{J}_{t} - \Big[{\mathbf{G}^{'}}^i \cdot \mathbf{J}_{in}^i \Big]_{i=1}^m,
\end{equation}
\begin{equation}
    \hat{\mathbf{L}}_t = \Big[[\mathbf{W} \cdot \mathbf{G}^{'}]^i \cdot \hat{\mathbf{L}}^i \Big]_{i=1}^l + \mathbf{W} \cdot \mathbf{O}.
    \label{eq:inv_L}
\end{equation}
An example is shown in Fig. \ref{fig:inv_res}(d). The linear transformation in Eq. \ref{eq:inv_L} does not provide a smooth unposed surface causing an offset error in some landmarks. Therefore, we propose an approximation to unpose the landmarks which is accurate and stable during training. Specifically, we compute unposed landmarks $\hat{\mathbf{L}}_t$ as\footnote{We discard $R(\phi)$ for simplicity of reading.}:

\begin{equation}
    \hat{\mathbf{L}}_t = \Upsilon^{-1}(\hat{\mathbf{L}}) + \mathbf{T}_t^*(\tilde{\rho}) - \Upsilon^{-1}(\Upsilon(\mathbf{T}_t^*(\tilde{\rho}))),
    \label{eq:inverse}
\end{equation}
where $\tilde{\rho}$ is the set of indices of the median vertex for each landmark patch and $\mathbf{T}_t^*(\tilde{\rho})=\{\mathbf{T}_t^{*i}:i\in\tilde{\rho}\}$. In Eq. \ref{eq:inverse} we update unposed landmarks ($\Upsilon^{-1}(\hat{\mathbf{L}})$) by summing to a correction offset. This offset is computed by the aid of a known reference body ($\mathbf{T}_t^*$ in this case). We apply a nested forward-backward kinematic function to $\mathbf{T}_t^*$ and subtract the results from the reference body. The result in Fig. \ref{fig:inv_res}(c) shows this is a valid approximation improving the unposing procedure.

\begin{table*}[!ht]
    \centering
    \begin{tabular}{|c|c|c|c|}
        \hline
        Method & $\mathbf{J}_{in}$ & $\mathbf{J}_{out}$ & $\mathbf{T}_{out}$ \\ \hline
        Preprocessing + \textit{DeepMurf} + $\Psi_1$ & 16.8 & 19.2 & 22.7 \\ \hline
        \textit{DeepMurf} + $\Psi_1$ (cascading) & 34.7 & 41.6 & 47.2 \\ \hline
        \textit{DeepMurf} + $\Psi_1$ - $\mathcal{L}_{unpose}$ (without inverse kinematic loss) & 97.6 & 42.1 & 47.6 \\ \hline
        \textit{DeepMurf} & 35.9 & 56.8 & 64.6 \\ \hline
        \textit{DeepMurf}, $\mathcal{L}_T$ with hard assignment & 34.4 & 61.7 & 71.1 \\ \hline
        \textit{DeepMurf} - $\mathcal{L}_J$ & 44.3 & 65.3 & 72.7 \\ \hline
        \textit{DeepMurf} - $\mathcal{L}_\phi$ & 42.6 & 63.1 & 77.3 \\ \hline
        \textit{DeepMurf} - $\mathcal{L}_\beta$ & 44.8 & 70.1 & 80.4 \\ \hline
        $\Psi_0$ + $\mathcal{L}_2$ - $\mathcal{L}_J$ - $\mathcal{L}_{unpose}$ (without attention model) & - & 85.3 & 92.9 \\ \hline \hline
        \textit{DeepMurf} (trained on 256 samples) & 50.8 & 102.9 & 121.5 \\ \hline
        \textit{DeepMurf} - $\mathcal{L}_{unpose}$ (trained on 256 samples) & 127.4 & 121.7 & 136.5 \\ \hline
    \end{tabular}
    \caption{Ablation results on SURREAL dataset. The errors are in millimeters. \textit{DeepMurf}=ATN + $\Psi_0$ + $\mathcal{L}_2$}
    \label{tab:ablation_all}
\end{table*}

\textbf{Training procedure.} We train the network incrementally. We first train DAE, ATN and $\Psi_0$ end-to-end by $\mathcal{L}_1$ loss:
\begin{equation}
    \mathcal{L}_1=\lambda_1\mathcal{L}_{DAE}+\lambda_2\mathcal{L}_{\beta}+\lambda_3\mathcal{L}_{\phi}+\lambda_4\mathcal{L}_{J}+\lambda_5\mathcal{L}_{T}+\lambda_6\mathcal{L}_{unpose},
\end{equation}
\begin{equation}
    \mathcal{L}_2=\lambda_2\mathcal{L}_{\beta}+\lambda_3\mathcal{L}_{\phi}+\lambda_4\mathcal{L}_{J}+\lambda_5\mathcal{L}_{T}+\lambda_6\mathcal{L}_{unpose},
\end{equation}
where $\{\lambda_i\}_{i=1}^6$ are balancing terms set empirically as 1, 0.1, 1, 0.1, 10 and 2, respectively. We then freeze DAE, ATN and $\Psi_0$, and train $\Psi_1$ by $\mathcal{L}_2$ loss. We freeze $\Psi_1$ and train next cascades likewise.

\section{Experiments}
In this section, we first describe training details and considered datasets for the experiments. Then, we provide an extensive analysis of the proposed architecture components and loss functions. Finally, we show proof-of-concept real applications of mocap to body surface reconstruction.

\subsection{Training details}
The code was implemented on Tensorflow and the model was trained on a TITAN Xp GPU. All networks were trained by Adam optimizer with learning rate 0.001 (and default optimizer parameters), from scratch with Xavier initializer, batch size 256 and dropout keeping probability 0.8. The network could converge in less than 6K training steps. Processing time took 1.02s in training for 1 step with batch size 256, and 0.41s and 0.013s in testing for 1 step with batch size 256 and 1, respectively.

\subsection{Datasets}
\textbf{SURREAL \cite{varol2017surreal}.} It is composed of 68K videos of rendered humans on top of fixed RGB background. This is a synthetic dataset of humans generated with SMPL model, thus containing exact annotations. We use this dataset for ablation study. The dataset contains millions of frames. In this paper, we randomly subsample 88K and 27K frames from the training and  validation set, respectively. This dataset does not provide landmarks. Therefore we create them artificially. We use the 67 landmarks defined in \cite{loper2014mosh} for this dataset. For each patch $\rho_i$ associated to $i-$th landmark, we select a random point on the patch surface and move it in the direction of its normal using a random distance in the range of $[8..10]$mm.

\textbf{MOSH-SSM \cite{mahmood2019amass}.} This is a recently published mocap dataset of around 4.5K frames captured from two females. Each frame has an accurate 3D scanned data. In overall, 73 landmarks have been used in the whole dataset (a subset of \cite{loper2014mosh} landmarks) and in average 10 landmarks are missing in each frame. This dataset has a small variability in pose. We are interested in this dataset due to its availability of synchronized scanned bodies and real-world mocap data challenges.

\textbf{CMU \cite{cmu2001}.} This is a widely used large mocap dataset captured from 96 subjects with more than 1.9K motions. This dataset contains 41 landmarks and the rate of missing landmarks is low.

\textbf{TCD Hands \cite{hoyet12hand}.} We use this dataset to analyze our approach in predicting expressive human body, i.e. body plus hands. There is just one subject in this dataset performing 62 motions. We use 46 standard landmarks on the body plus 8 landmarks on each hand (proposed in \cite{hoyet12hand}, i.e. 4 landmarks for thumb and 4 fingertips). 

Each standard landmark has an alphabetical code to recognize it. We define a dictionary of landmark codes and their corresponding SMPL vertex indices for each patch. This way we can easily switch between datasets as long as landmark codes follow the standard labels.

\begin{figure*}
    \centering
    \includegraphics[width=.9\textwidth]{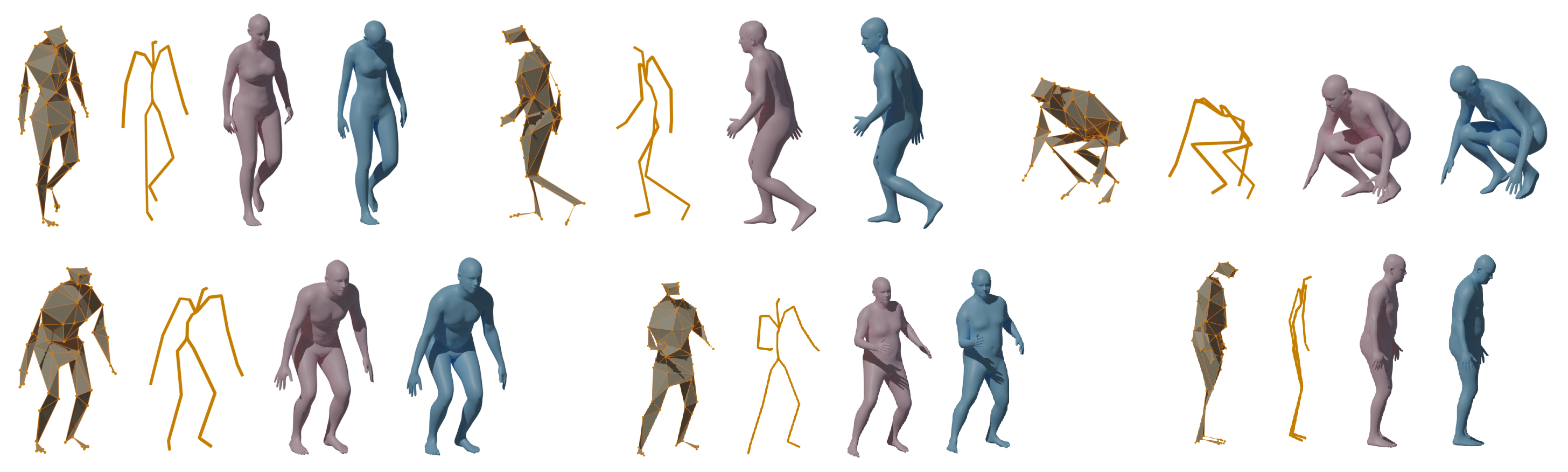}
    \caption{This figure shows the different stages of the cascading model for some samples of SURREAL dataset. First, we see the input landmarks. Next, the estimated joints obtained through ATN. Finally, the estimated 3D human model (pink) along the ground truth (blue).}
    \label{fig:q_surreal}
\end{figure*}

\begin{table*}[!ht]
    \centering
    \begin{tabular}{|c|c|c|c|c|c|}
        \hline
        Method & $\tau$ & DAE & $\mathbf{J}_{in}$ & $\mathbf{J}_{out}$ & $\mathbf{T}_{out}$ \\ \hline
        Preprocessing + \textit{DeepMurf} + $\Psi_1$ & 0 & - & 16.8 & 19.2 & 22.7 \\ \hline
        \parbox[t]{70mm}{\multirow{3}{*}{Preprocessing + DAE + \textit{DeepMurf} + $\Psi_1$ + $\mathcal{L}_{DAE}$} }
        & 0.1 & 8.1 & 23.7 & 28.2 & 33.2 \\ \cline{2-6}
         & 0.3 & 7.8 & 33.3 & 33.6 & 41.5 \\ \cline{2-6}
         & 0.5 & 6.8 & 37.4 & 39.4 & 45.7 \\ \hline 
        \textit{DeepMurf} & 0 & - & 35.9 & 56.8 & 64.6 \\ \hline 
        \parbox[t]{70mm}{\multirow{3}{*}{DAE + \textit{DeepMurf} + $\mathcal{L}_{DAE}$} }
        & 0.1 & 11.2 & 48.4 & 59.7 & 68.3 \\ \cline{2-6}
         & 0.3 & 11.9 & 69.7 & 77.1 & 87.5 \\ \cline{2-6}
         & 0.5 & 12.4 & 85.3 & 87.5 & 100.4 \\ \hline 
    \end{tabular}
    \caption{The impact of training with missing landmarks on SURREAL validation set. $\tau$ is the rate of missing landmarks.}
    \label{tab:missing_landmarks}
\end{table*}

\begin{figure}[!ht]
    \centering
    \includegraphics[width=\linewidth]{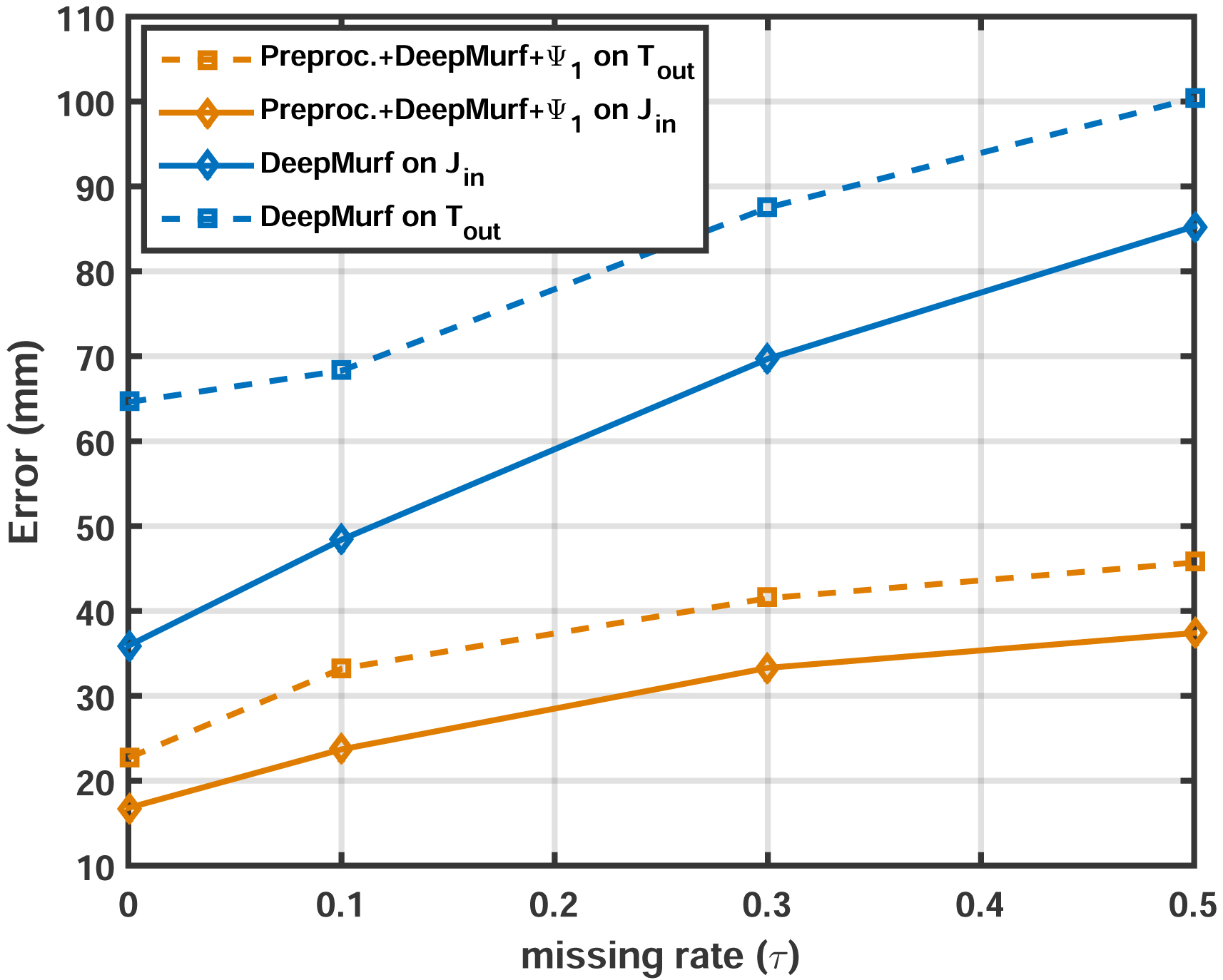}
    \caption{The impact of training with missing landmarks on SURREAL validation set.}
    \label{fig:missing_rate}
\end{figure}

\subsection{Results}
In this section, we study ablative results of our proposed approach on SURREAL validation set. We also show how our method performs on a real world scenario as in MOSH-SSM dataset and we compare to \cite{loper2014mosh,mahmood2019amass} on this dataset. Finally, we show qualitative results on both datasets.

\subsubsection{Ablation study}
Our base model is ATN + $\Psi_0$ trained with $\mathcal{L}_2$. We call this model \textit{DeepMurf}. We then explain the results by adding or removing different components to/from \textit{DeepMurf}. To evaluate on SURREAL dataset, we report average per joint/vertex Euclidean error on $\mathbf{J}_{in}$, $\mathbf{J}_{out}$ and $\mathbf{T}_{out}$ in millimeters. The results are shown in Tab. \ref{tab:ablation_all} and \ref{tab:missing_landmarks}. 

\textbf{Impact of attention model ATN.} Attention model brings several advantages: 1) an accurate estimation of joints $\mathbf{J}_{in}$ (as in \textit{DeepMurf} with an error of 35.9mm), and 2) applicability of additional loss functions, i.e. $\mathcal{L}_J$ and $\mathcal{L}_{unpose}$. By omitting $\mathcal{L}_J$ from \textit{DeepMurf} training, one can observe that the surface error is increased by 8mm (6th row in Tab. \ref{tab:ablation_all}). As an additional experiment, we omit ATN, and consequently $\mathcal{L}_J$ and $\mathcal{L}_{unpose}$, from \textit{DeepMurf} and train $\Psi_0$ directly by feeding landmarks $\hat{\mathbf{L}}$ to the network. As a result, surface error is increased by more than 28mm (9th row in Tab. \ref{tab:ablation_all}). This shows that the proposed ATN has a huge impact on the results and \textit{DeepMurf} without ATN is not able to properly learn useful information just from landmarks to map them to the pose and shape parameters.

\textbf{Impact of regularization loss on pose and shape parameters.} We omit $\mathcal{L}_\phi$ or $\mathcal{L}_\beta$ from $\mathcal{L}_2$ loss and train \textit{DeepMurf}. As a result (7th and 8th rows in Tab. \ref{tab:ablation_all}), the error is increased by around 13mm and 16mm for $\mathcal{L}_\phi$ and $\mathcal{L}_\beta$, respectively. This is mainly due to the sensitivity of SMPL to the noise and convergence to invalid parameters in the backpropagation. As one can see, omitting $\mathcal{L}_\beta$ has more impact on the results than $\mathcal{L}_\phi$.

\textbf{Soft vs. hard landmark-to-surface assignment.} \textit{DeepMurf} has a surface error of 64.6mm. When \textit{DeepMurf} is trained with a hard landmark-to-surface assignment in $\mathcal{L}_T$, the surface error is increased by more than 6mm (5th row in Tab. \ref{tab:ablation_all}). A hard assignment means each landmark is always associated with a fixed vertex on the SMPL surface. A hard assignment introduces some noise in the loss and does not lead to the most optimum solution. We also applied a chamfer distance as $\mathcal{L}_T$. However, it had a high complexity and did not converge well. These results reveal that our soft assignment works well in practice and can cope with the challenges in the data.

\begin{figure*}[!ht]
    \centering
    \includegraphics[width=0.9\textwidth]{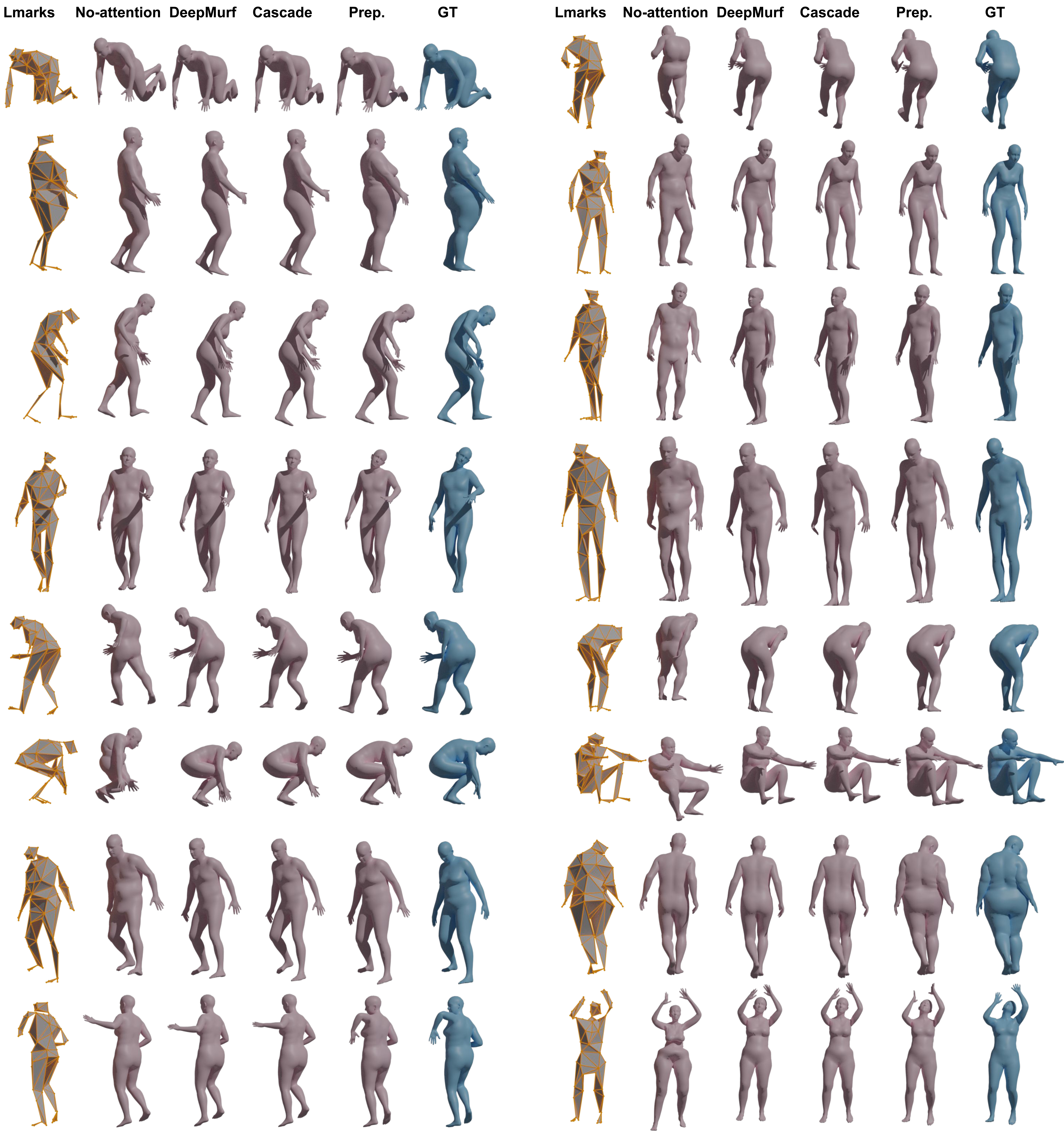}    
    \caption{Qualitative ablation results on SURREAL validation dataset. Connections are added to the landmarks for visualization purposes.}
    \label{fig:q_ablation}
\end{figure*}

\textbf{Impact of cascading.} In our proposed cascading, each block learns the error of the previous block. As one can see (2nd row in Tab. \ref{tab:ablation_all}), an additional cascading block $\Psi_1$ to \textit{DeepMurf} improves the surface error by more than 17mm. We have observed more cascading blocks were not as effective as $\Psi_1$ improving the error around 1mm. We note that \textit{incrementally} training cascading approach is important for performance gains. Training cascading network end-to-end from scratch performs similar to \textit{DeepMurf}. We show some qualitative images of cascading model on SURREAL dataset in Fig. \ref{fig:q_surreal}.

\textbf{Impact of inverse kinematic loss.} We study inverse kinematic loss $\mathcal{L}_{unpose}$ in two ways. Firstly, we omit $\mathcal{L}_{unpose}$ from \textit{DeepMurf}+$\Psi_1$ network\footnote{Note that \textit{DeepMurf} - $\mathcal{L}_{unpose}$ behaves similarly.}. As it is visible in Tab. \ref{tab:ablation_all} (3rd row), $\mathcal{L}_{unpose}$ has a huge impact on $\mathbf{J}_{in}$. Omitting $\mathcal{L}_{unpose}$ from cascading network increases $\mathbf{J}_{in}$ error by around 63mm reducing performance of ATN. However, $\mathcal{L}_{unpose}$ does not have much impact on the surface error. Secondly, we omit $\mathcal{L}_{unpose}$ from \textit{DeepMurf} and train the model for 500 epochs on a very small dataset (256 samples). We want to study generalization ability of the proposed loss on small custom datasets. As a result (can be seen in the last two rows), $\mathcal{L}_{unpose}$ helps to gain 15mm improvement on the surface error. Interestingly, \textit{DeepMurf} can still perform well to estimate $\mathbf{J}_{in}$ (error of 50.8mm) trained on such a small dataset.

\begin{figure*}
    \centering
    \includegraphics[width=.9\textwidth]{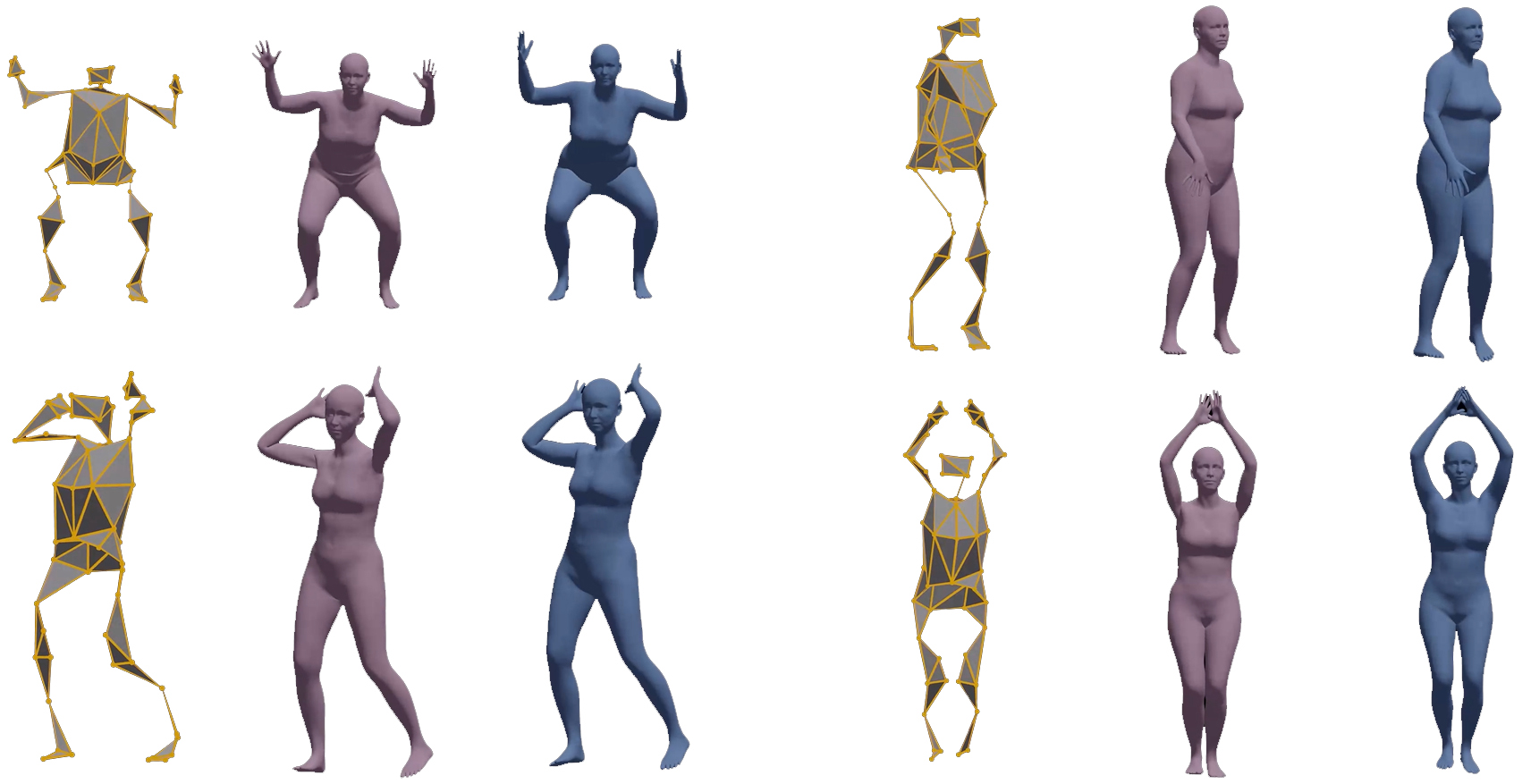}
    \caption{Qualitative comparison with Mosh++ \cite{mahmood2019amass} on MOSH-SSM dataset. Left: input landmarks, middle: our predictions with 67 landmarks, and right: Mosh++ predictions.}
    \label{fig:q_mosh}
\end{figure*}

\textbf{Impact of global orientation.} Data normalization and augmentation are two common preprocessing techniques applied in deep learning to boost performance. In this paper we mainly focus on the data normalization. In the previous experiments, we applied a translation invariant solution by subtracting the input landmarks from the mean point. Here, we explore an additional preprocessing to make the network rotation invariant. To do so, we rigidly (without scaling) align all landmarks in the dataset to a reference set of landmarks, e.g. the template landmarks. More specifically, we apply procrustes analysis to compute a rotation matrix and translation vector to transform landmarks $\hat{\mathbf{L}}$. We then train the cascading network on the aligned data as before. At test time, estimated surface is transformed back to the original orientation. The results are shown in the first row of Tab. \ref{tab:ablation_all}. As it can be seen in the comparison of the first two rows, the surface error is reduced by 48\% (24.5mm).

\textbf{Impact of missing landmarks.} To study the impact of missing landmarks, we train DAE+\textit{DeepMurf} and Preprocessing+DAE+\textit{DeepMurf}+$\Psi_1$ (including $\mathcal{L}_{DAE}$ in the training) with different rates of missing landmarks, that is, we randomly select 90\%, 70\% and 50\% of the landmarks in the batch and assign zero to the rest. This is repeated for each step. This means in average 7, 20 and 33 landmarks are dropped for each frame in each setup. The results can be seen in Tab. \ref{tab:missing_landmarks}. Interestingly, DAE error on missing landmarks (3rd column) is not strongly correlated to the rate of missing landmarks ($\tau$). This error is around 7.5mm and 11.8mm in average for the cascading and baseline models, respectively. Also, the error on the surface ($\mathbf{T}_{out}$) and the input joints ($\mathbf{J}_{in}$) is polynomial with a degree below 1 w.r.t. the $\tau$. This means the error will not increase much in higher rates of missing landmarks. This can be seen in Fig. \ref{fig:missing_rate}. We note that the model trained with preprocessing is more resistant against the missing landmarks than the default model \textit{DeepMurf}. These results show that DAE is effective against missing landmarks to be used in real world applications of the proposed surface recovery.

\textbf{Qualitative comparison.} We qualitatively compare different methods in the ablation study in Fig. \ref{fig:q_ablation}. As expected, according to Tab. \ref{tab:ablation_all}, the attention model has a high impact on the quality of the results. Interestingly, body shape has more impact on the error than pose, e.g. in the extreme shapes. Finally, by applying the preprocessing, we can generate a near perfect body surface.

\begin{figure*}
    \centering
    \includegraphics[width=.9\textwidth]{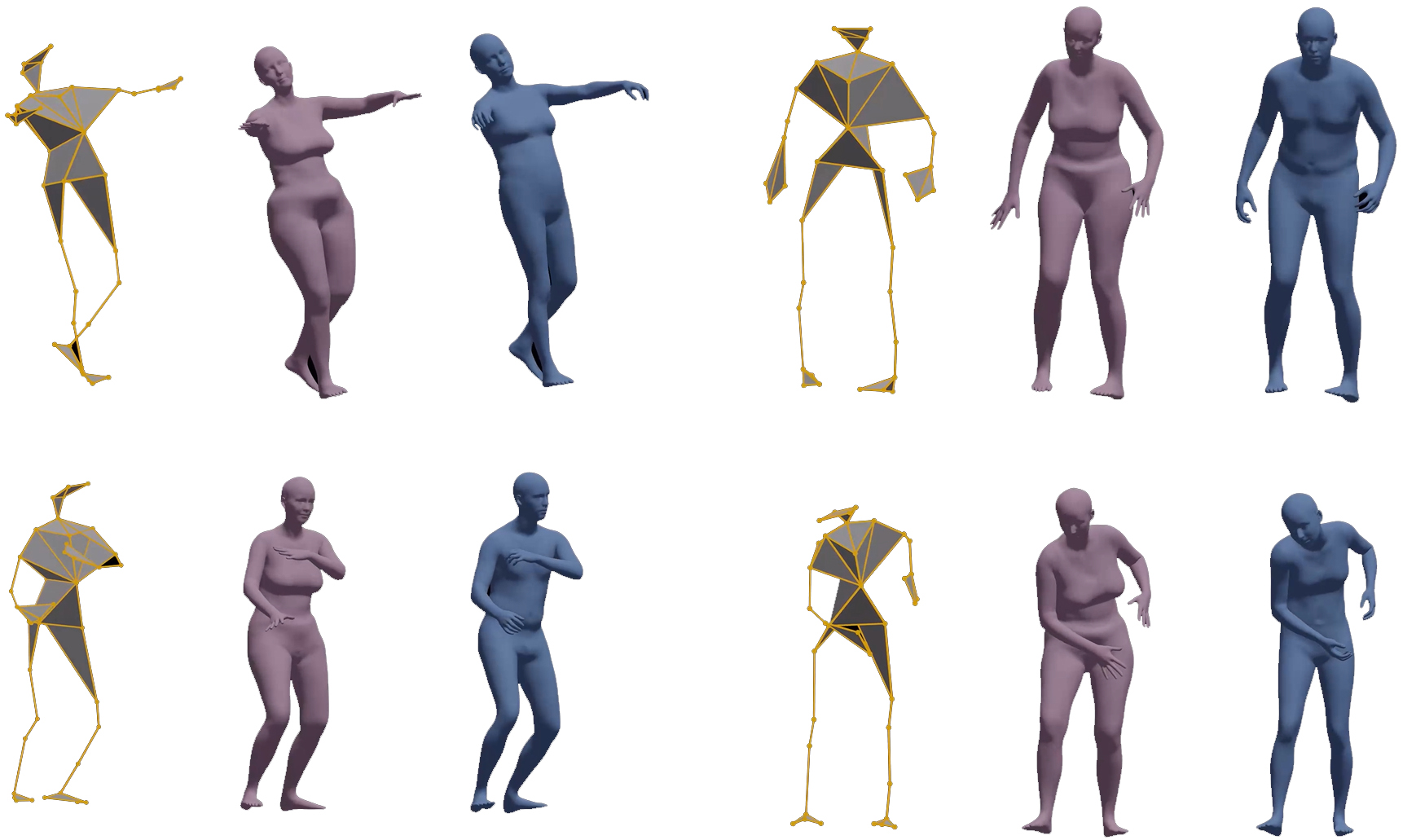}
    \caption{Qualitative comparison with Mosh++ \cite{mahmood2019amass} on CMU dataset. Left: input landmarks, middle: our predictions, and right: Mosh++ predictions.}
    \label{fig:q_cmu}
\end{figure*}

\subsubsection{MOSH-SSM results}
This dataset does not have any split regarding training-testing set. Therefore, we randomly split the data in 50/50\% ratio and train our cascading model DAE + \textit{DeepMurf} + $\Psi_1$ (including preprocessing). To evaluate our approach on this dataset, we compute average scan-to-model distance as in \cite{mahmood2019amass}. To do so, for each frame we randomly sample 10K points from the ground truth scan and for each point take its nearest neighbor vertex on the estimated $\mathbf{T}_{out}$. Finally, per point Euclidean distances are averaged over the whole dataset. The scans and landmarks are not very well aligned in this dataset. Therefore, to apply scan-to-model distance we first align predictions with scans by fitting a translation vector through CPD algorithm \cite{myronenko2010point}. 

\begin{table}[!ht]
    \centering
    \begin{tabular}{|c|c|}
        \hline
        Method & scan-to-model error \\ 
         & (mm) \\ \hline
        Mosh++ \cite{mahmood2019amass} & 18.1 \\ \hline 
        Best \textit{DeepMurf} (46 landmarks) & 24.5 \\ \hline 
        Best \textit{DeepMurf} (67 landmarks) & 19.9 \\ \hline 
        Best \textit{DeepMurf} (67 landmarks) &  \\
        + & 19.8 \\
        temporal smoothing &  \\ \hline 
    \end{tabular}
    \caption{Quantitative results on Mosh-SSM dataset.}
    \label{tab:ssm}
\end{table}

We evaluate three different models on this dataset and compare with Mosh++ \cite{mahmood2019amass} in Tab. \ref{tab:ssm}. In the second row, we train the model with a set of 46 standard landmarks. This model has the highest error among others (24.5mm) due to a reduced set of landmarks. However, it still performs well. In the next row in the table, we train the network with 67 landmarks (as in \cite{mahmood2019amass}) which shows more than 21\% improvement against 46 landmarks. This is while the difference error with Mosh++ is 1.8mm. This model is trained and tested on single frames and temporal smoothing is not applied. In the next experiment we apply temporal smoothing as a post-processing. To do so, we set shape parameters as the average over the whole sequence. Regarding the pose parameters, we check for jittering based on angle difference between previous and next frames for each joint and axis. We empirically select threshold $0.1$ for this task. As a result, the error is improved by 0.1mm which shows predictions on single frames are temporally consistent. Finally, we show qualitative results in Fig. \ref{fig:q_mosh}.

\subsubsection{CMU results}
We analyze our approach on CMU dataset qualitatively in Fig. \ref{fig:q_cmu}. As one can see, we predict similar surfaces to Mosh++ on this dataset with variable motion and subjects. We achieve this performance in few miliseconds vs several minutes of Mosh++. Since we train a neutral SMPL body, without specifically defining gender, our model converges mostly to a female body using standard landmarks. Similar to Mosh++, one can train gender specific SMPL models when gender specific landmarks are not available.

\begin{figure*}
    \centering
    \includegraphics[width=.9\textwidth]{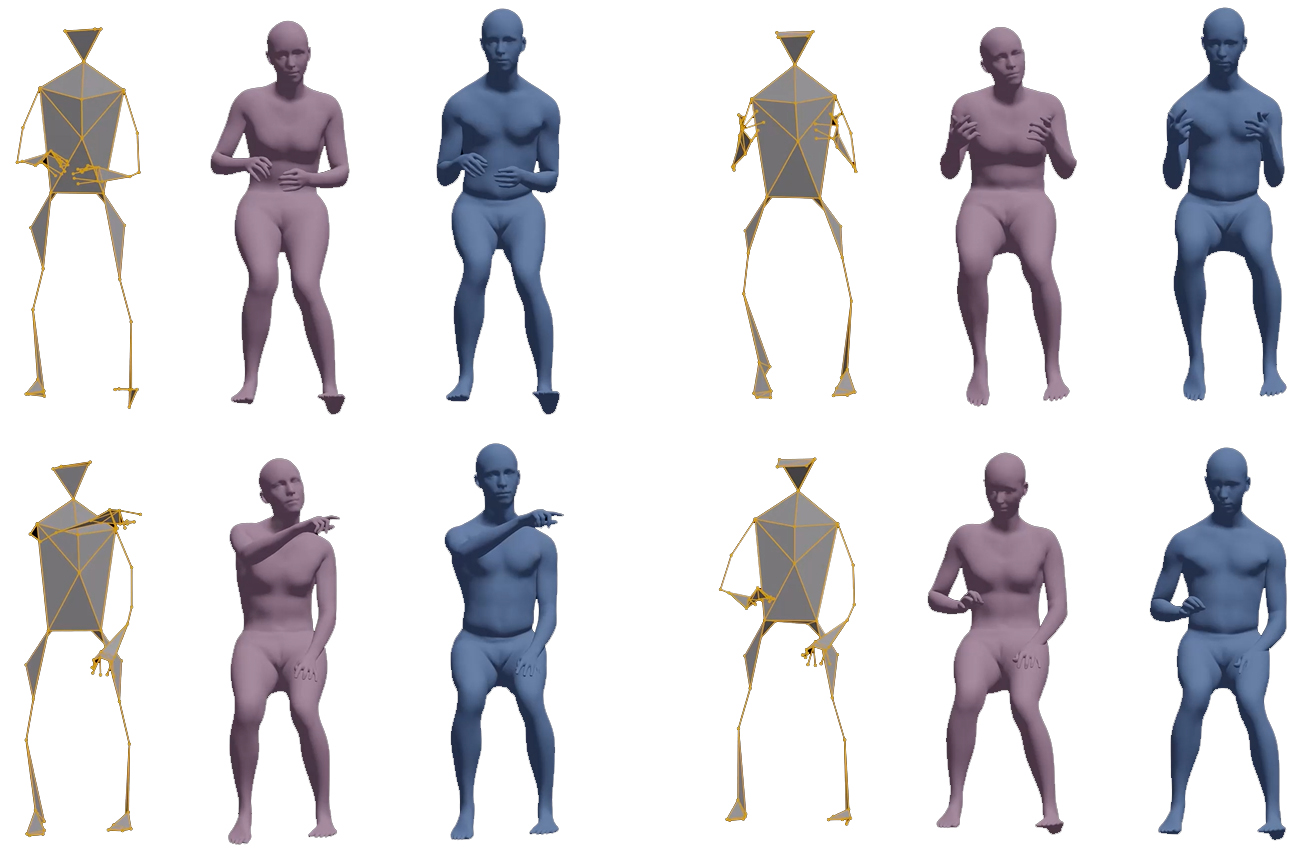}
    \caption{Qualitative comparison with Mosh++ \cite{mahmood2019amass} on TCD Hands dataset. Left: input landmarks, middle: our predictions, and right: Mosh++ predictions.}
    \label{fig:q_tcd}
\end{figure*}

\subsubsection{TCD Hands results}
In this dataset, we analyze the applicability of our approach for the case of expressive humans, specifically body and hands. Handling body and hand pose together in a single network is a challenging task due to unbalanced landmarks, different motion and level of details between body and hands. To do so, we update \textit{DeepMurf} with SMPLH model \cite{mahmood2019amass}. We train the network by setting high weights on lower arm joints and landmarks in $\mathcal{L}_{\phi}$ and $\mathcal{L}_T$, respectively. We train the network with the same capacity as before. The results can be seen in Fig. \ref{fig:q_tcd}. As it can be seen, our approach shows promising results being applicable for expressive humans.

\subsection{Applications}
Due to the popularity of mocap data in movie editing and videogame industry, a fast, accurate simulation and rendering can save a lot of working hours of animators. Furthermore, the deep model can be embedded into graphic engines for specific applications. In this section, we propose different applications for our deep-based mocap-to-surface estimation, shown in Fig. \ref{fig:motivation}. A basic application can be shape or pose modification. We also perform garment simulation on top of the reconstructed body. Furthermore, we apply retargeting to a bunny avatar. This can be done by replacing rigged SMPL template by any other rigged template consistent with SMPL inline functionality or replacing SMPL with any other generative model. All of this is done by just inputing a set of sparse landmarks to the application.

\section{Conclusions}
We presented a deep unsupervised approach for estimation of body surface from sparse mocap data. We applied a denoising autoencoder network able to recover missing landmarks accurately. Our proposed attention model estimated body joints from landmarks and we showed it has a high impact on the accuracy of the generated surface. Attention model also allowed us to apply several loss functions improving the model performance including an unposing layer useful to learn body geometry. We also designed a cascading regression which helped to improve the error by 17mm. Our quantitative and qualitative results on four datasets show applicability of our approach in real world problems (including expressive humans) with a surface error less than 20mm.

Although we showed promising results for expressive humans, there is still room for improvement. Also, we did not model soft-tissue on \textit{DeepMurf}, which can be an important source of landmarks noise. Body soft-tissue dynamics can be modeled from body pose and shape. We will explore these ideas as the future work.

\section*{Acknowledgements}
This work is partially supported by ICREA under the ICREA Academia programme, and by the Spanish project PID2019-105093GB-I00 (MINECO / FEDER, UE) and CERCA Programme / Generalitat de Catalunya, and by Amazon Research Awards ARA.

\bibliographystyle{ieee}
\bibliography{main_arxiv}
\end{document}